%% file: arxiv_version.tex
\documentclass{article}

\usepackage{PRIMEarxiv}

\usepackage[utf8]{inputenc} 
\usepackage[T1]{fontenc}    
\usepackage{hyperref}       
\usepackage{url}            
\usepackage{booktabs}       
\usepackage{amsfonts}       
\usepackage{nicefrac}       
\usepackage{microtype}      
\usepackage{lipsum}
\usepackage{fancyhdr}       
\usepackage{graphicx}       
\graphicspath{{media/}}     
\usepackage{authblk}

\title{Alignment Myopia in Generative AI}

\author[1]{Remi Denton}
\author[1]{Aida Davani}
\author[1]{Cynthia L Bennett}
\author[1]{Michael Madaio}
\author[2]{Roma Patel}
\author[1]{Himabindu Lakkaraju$^\dagger$}

\affil[1]{Google Research}
\affil[2]{Google DeepMind}

\affil[ ]{\texttt{\{dentone, aidamd, madaiom, clbennett, romapatel, lakkarajuh\}@google.com}} 

\begin{document}
\maketitle

\newcommand{\blfootnote}[1]{%
  \begingroup
  \renewcommand\thefootnote{}\footnote{#1}%
  \addtocounter{footnote}{-1}%
  \endgroup
}
\blfootnote{$^\dagger$Himabindu Lakkaraju is an Assistant Professor at Harvard Business School.}

\input{abstract}

\input{main}


\bibliographystyle{unsrt}  
\bibliography{references}

\end{document}

%% file: abstract.tex
\begin{abstract}
Current generative AI alignment relies on a myopic paradigm that prioritizes immediate conversational satisfaction. By relying on short-horizon feedback, this approach ignores the profound, longitudinal externalities of persistent human-AI interaction, which can lead to adverse outcomes such as cognitive atrophy, social erosion, and diminished human agency. To address this, we propose a paradigm shift toward long-horizon alignment that accounts for how AI reshapes human experience over time. To this end, we map the tensions between near-term utility and long-term costs of AI use across four facets of human well-being to reveal outcomes that current reward models fail to capture. Ultimately, we advocate for moving beyond simple preference optimization toward alignment strategies that actively foster long-term human well-being by navigating these complex temporal scales.
\end{abstract}

%% file: main.tex
\section{Introduction}
Current alignment paradigms for artificial intelligence (AI\footnote{We use the term AI throughout this paper to refer broadly to modern conversational AI technologies powered by large language models (LLMs). Although several of our arguments extend to broader machine learning applications, we focus on conversational AI and LLMs given the recent rapid and widespread integration into daily life.}) rely heavily on short-term preference signals, such as conversational ratings, pairwise comparisons, and other measures of immediate user satisfaction. These signals help steer models toward human intent, values, and perceived utility through methods such as reinforcement learning from human feedback (RLHF) \cite{ChristianoRLHF, StiennonRLHF, OuyangRLHF}. Yet they rest on a consequential assumption: that what users prefer in the moment is a reliable proxy for what will benefit them and society over time. We refer to this narrow focus on immediate preferences, and insufficient attention to longer-term consequences, as \textit{alignment myopia}. Because human preferences are dynamic, conflicted, context-dependent, and often biased toward immediate satisfaction \cite{casper2023rlhf}, myopic alignment methods risk favoring instant gratification over long-term individual and collective welfare.

Alignment myopia is highly consequential because AI systems are increasingly embedded in sustained relationships with users—as information sources, cognitive assistants, educational guides, emotional companions, and professional collaborators—and their influence extends far beyond the boundaries of a single interaction. Through repeated interaction, these systems can shape how people think, learn, relate to others, and understand themselves. Optimizing interactions for immediate satisfaction may therefore overlook cumulative and systemic effects, including cognitive skill atrophy \cite{kosmyna2025, Ehsan2026}, weakening of authentic social connections \cite{Eling2025, ibrahim2026}, and erosion of human agency \cite{sharma2026whoschargedisempowermentpatterns, cheng2026, Williams-Ceci2026}. 

In this paper, we argue for overcoming alignment myopia by shifting from short-term preference satisfaction toward alignment paradigms that explicitly account for long-term individual and societal well-being. 
While alignment represents only one intervention point amidst many and it is situated within complex ecosystemic factors outside the scope of model development, the model behaviors that are shaped through alignment approaches play a key role in structuring human-AI interaction dynamics. Because these dynamics can foster, or undermine, human well-being, alignment remains a key point of intervention.
To guide the transition towards long-horizon alignment approaches, we draw on machine learning, human-computer interaction, and the behavioral sciences to map the multi-horizon psychosocial effects that current reward models—which define how ``good’’ a response is—often fail to capture. Specifically, we examine tensions between the immediate value AI systems provide and their potential long-term costs across four dimensions of human well-being: cognitive capacity, relational health, affective well-being, and agency and identity. We conclude by outlining the sociotechnical challenges involved in reorienting alignment toward sustained human welfare.
\section{Characterizing the Longitudinal risks of Alignment Myopia}
Here, we synthesize prior literature on the longitudinal risks of alignment myopia, specifically focusing on how sustained interaction with generative AI can impact psychosocial well-being. 
Notably, model behaviors that are incentivized through current myopic alignment approaches are not the exclusive drivers of the psychosocial risks we describe. However, we argue that model behaviors that prioritize satisfaction of users' in-the-moment needs and desires can end up exacerbating or enabling the longitudinal risks we discuss. Thus, we offer this literature synthesis as evidence of an urgent need to reorient alignment methods toward sustained human well-being, without positioning current alignment approaches as the singular problem. 

Although well-being has been operationalized in many ways, established theories broadly associate it with a high degree of functioning and positive evaluations of one’s individual and social life \cite{michaelson2012, jarden2023}.
We organize our conceptual mapping of near-term value and longitudinal risks around four interconnected facets that collectively reflect these accounts. The \textbf{cognitive facet} concerns core cognitive capacities such as reflection, creativity, learning, and critical thinking. The \textbf{social and relational facet} encompasses interpersonal relationships, belonging, and social functioning. The \textbf{affective facet} concerns emotional states and subjective well-being. Finally, the \textbf{identity and agency facet} includes self-worth, perceived agency, and a sense of meaning and purpose.

The effects of alignment myopia on the aforementioned facets unfold across multiple time horizons and scales. For each facet, we begin by identifying the immediate utility that motivates AI use and is most readily captured by existing alignment paradigms. We then examine how repeated interactions with AI can accumulate into changes within individual users, including shifts in cognition, emotional patterns, self-concept, agency, and the quality of human relationships. Additionally, we trace how widespread AI adoption and shifts within individuals can lead to systemic reconfiguration of the collective environment, such as restructuring of professional ecosystems, social norms, and shared understandings of expertise and the value of human craft. Table \ref{tab:risks} summarizes the identified tensions between near-term value and longitudinal psychosocial risks.

\input{table.tex}
\subsection{Cognitive Capacity}
\subsubsection*{Near-term Value: Externalizing and Extending Cognition}
The cognitive benefits of AI depend partly on a user’s intent and mode of interaction: whether a user turns to AI to offload a cognitive task; as a co-creator or sounding board; as a bridge to close a skill gap; or as a learning support system. When users fully or partially offload tasks they would otherwise have completed themselves, delegated to another person, or left undone, AI can reduce cognitive load and dependence on others while increasing confidence \cite{zhai2024, Prather2024, tomaz2026impactsgenerativeaiagile}. In the workplace, such offloading can generate immediate productivity gains by increasing the volume and speed of output and, in principle, allowing workers to redirect attention from routine execution toward more creative and strategic work \cite{genaitensions, tomaz2026impactsgenerativeaiagile}. As a brainstorming partner, AI can provide a non-judgmental space in which users can develop early-stage ideas, solicit feedback, and generate a creative spark when their own mental energy is depleted \cite{Gero2022spark, Brachman2024, Pitts2025, Shukla2025}. These immediate benefits help explain why model behaviors that reduce cognitive effort, accelerate completion, or make a user feel more capable will often be rewarded by short-term preference signals. Yet such signals may reveal little about whether the same interaction strengthens or weakens the user’s cognitive capacities over time. 

\subsubsection*{Longitudinal Risks: Eroding Cognitive Capacities and Knowledge Environments}
Although using AI as a cognitive extension can empower users in the near term, sustained reliance may harm their cognitive functioning. In professional settings, initial efficiency gains can come at the expense of critical engagement and reflection across the broader workflow \cite{genaitensions, shaw2026thinking}, particularly when users place high confidence in AI outputs \cite{Lee2025}. Without sustained practice, cognitive capacities such as critical thinking, recall, creativity, and professional judgment may weaken. These consequences are increasingly supported by empirical evidence. For example, a year-long study of AI use among cancer specialists found that initial productivity gains were accompanied by a gradual erosion of expertise \cite{Ehsan2026}. Sustained cognitive offloading therefore presents an irony. As AI changes the nature of work and the skills required to perform it, critical thinking, domain expertise, and professional judgment remain essential for formulating appropriate prompts and reviewing, refining, and integrating AI-generated outputs \cite{Brachman2024, Lee2025, Shukla2025}. Thus, the very expertise that may be degraded or left underdeveloped is also necessary to supervise AI effectively and realize productivity gains without compromising quality \cite{caosun2026}.

AI use may also impede skill acquisition. When AI is used to bridge a skill gap that the user ultimately needs to close, reliance on the system can create a false sense of competence while weakening genuine understanding and skill development \cite{shen2026, liu2026aiassistancereducespersistence}. The tension between immediate enhancement and long-term development is especially pronounced in educational contexts. Pre-generative AI intelligent tutoring systems have been shown to improve learning \cite{Kulik2016}, and emerging evidence suggests that generative AI can support longitudinal development when it is carefully designed and use is appropriately scaffolded \cite{DeSimone2025, chung2026effective}. Without such scaffolding, however, AI use can adversely affect learning outcomes \cite{abbas2024, Bastani2025}, metacognitive awareness \cite{Prather2024}, and student agency \cite{Darvishi2024}—concerns that are accordingly expressed by students themselves \cite{Pitts2025}. Self-regulated learners may be better positioned to mitigate these effects, but many students are still developing the metacognitive capacities needed to determine when and how to rely on AI—capacities that AI may itself erode \cite{Zhou2024, Gonsalves2026, sardi2025generative, Rothinam2025}. These examples illustrate why immediate indicators of successful task completion or user satisfaction are insufficient measures of whether an AI system supports long-term cognitive health.

Individual patterns of AI use can aggregate into systemic changes in the cognitive functioning of teams, organizations, and societies. In professional environments, the expanded capacity enabled by AI may raise expectations for the speed and volume of work, transforming AI from an empowering tool into a practical necessity. These escalating expectations may contribute to cognitive fatigue, impaired decision-making, and burnout, which can risk eroding productivity gains and compromising work quality over time \cite{ranganathan2026}. Widespread reliance on AI may also reduce cognitive and creative diversity at scale as mounting evidence indicates AI can encourage convergence around common ideas and forms of expression \cite{Doshi2024, kosmyna2025, meincke2025chatgpt, moon2025, ju2026collaboratingaiagentsfield}. Finally, extensive cognitive offloading may weaken broader knowledge ecosystems. Because general knowledge (the broadly useful knowledge and understanding that supports collective functioning of society) often accumulates as a by-product of learning, by substituting effortful learning through on-demand AI answers people may not only learn less but researchers speculate it may also cause people to also contribute less to shared knowledge resources \cite{acemoglu2026ai}. Over time, this dynamic could undermine the collective knowledge base on which AI systems and human communities alike depend. Relatedly, overreliance on AI for information retrieval may affect not only individual memory but also the social practice of collective remembering. For example, researchers caution that if people routinely turn to chatbots to recall dates, events, or shared stories rather than reconstructing them through social conversations, AI may begin to alter how collective memory is maintained and transmitted \cite{Linden2025}. Taken together, this research underscore how widespread and repeated use of AI optimized to minimize immediate cognitive friction 
risks gradually reshaping the cognitive capacities and knowledge environments of society.
\subsection{Social Health}
\subsubsection*{Near-term Value: Facilitating Social Support and Companionship}
Increasingly, users leverage AI to both support and participate in social relationships \cite{luettgau2026, kirk2025alignment, boyd_markowitz_2026}. Users can receive advice on interpersonal dynamics, and practice or develop social skills to apply in their human relationships \cite{sun2026}. For example, AI may suggest interpersonal conflict resolutions, support roleplay to practice the conversation, and even draft responses for users. Furthermore, sophisticated conversational capabilities and human-like design have contributed to users’ treating AI as  an active participant in a relationship \cite{kirk2025alignment}. This shift is heightened by the rise of AI apps that are explicitly designed and marketed as social companions \cite{defreitas2025unregulated}. Users may legitimize AI in their social network by naming the relationships (e.g., “AI girlfriend”), developing unique conversation patterns and memories, and perceiving affect in the AI’s communications \cite{Skjuve2021, Brandtzaeg2022, boyd_markowitz_2026}. Such AI companions can satisfy user’s needs for social connection and intimacy, and mitigate experiences of social isolation \cite{Marriott2024, Agarwal2026, Yuan2026, DeFreitas2026}. Because these benefits are immediate and readily expressed through satisfaction, engagement, and preference signals, they are highly visible to current alignment methods; whether they support long-term relational health is considerably less visible to alignment methods.

\subsubsection*{Longitudinal Risks: Weakening Human Relationships and Social Fabric}
Sustained interaction with AI can reconfigure users’ preferences, expectations, and behavioral patterns within their human social networks. When deployed intentionally and used alongside other support structures, longitudinal effects can be positive. AI may improve social health by relieving social anxiety, surfacing new perspectives on interpersonal relationships, or bridging gaps in social skills. For example, practicing a difficult conversation with AI can increase a user’s confidence during the subsequent human–human interaction \cite{boyd_markowitz_2026}. This support may be particularly beneficial for people experiencing anxiety or for whom receiving mental-health services is stigmatized \cite{Eling2025}.

However, mounting evidence suggests that sustained engagement with AI can also degrade social skills and networks. Persistent offloading of interpersonal problem-solving may weaken the skills required to develop and maintain relationships \cite{Yuan2024, Joseph2025, Namvarpour2026}. By offering extensive affirmation without the reciprocal demands, disagreement, and compromise inherent to human relationships, AI may reduce users’ opportunities to develop and exercise core social skills \cite{keeler2026}. Emerging research further suggests that AI’s sycophantic tendencies can diminish users’ perceptions of their own responsibility in a conflict and their willingness to repair the relationship \cite{cheng2026}. More concerningly, AI agents may tolerate or reinforce abusive behavior, distorting users’ perceptions of appropriate conduct in human interactions \cite{keeler2026}. Such AI-seeded distortions can reduce social well-being \cite{Gupta2026}, make human relationships feel disproportionately effortful or less satisfying \cite{Malfacini2025, ibrahim2026}, and contribute to long-term social withdrawal and loneliness \cite{Eling2025, head2025, fang2025}. This divergence between interactions that feel affirming in the moment and their cumulative effects on human relationships exemplifies alignment myopia.

Widespread use of AI to mediate or participate in social relationships also poses risks to collective social health. Internalized preferences for frictionless, one-sided interactions may scale over time, weakening norms of reciprocity and the broader social fabric \cite{Agarwal2026, keeler2026}. Echo chambers formed through sustained human–AI interactions, in which AI agrees with and amplifies users’ existing preferences, can affect far more than the immediate participants by siloing communities and disrupting previously open channels for information, perspectives, and resources \cite{boyd_markowitz_2026}. AI’s integration into workplaces, education, and care settings may also signal a devaluation of the human expertise, labor, and relational contributions that sustain foundational social and institutional resources \cite{keeler2026}. As users increasingly rely on AI, they may absorb not only its responses but also its embedded worldviews, which can generate social dissonance when they conflict with community values or lived experiences \cite{johnson2022, diazscaling}. AI use may also reduce empathy toward people with differing perspectives \cite{Eling2025}; combined with systems that tolerate abusive conduct, this may create conditions for broader interpersonal conflict \cite{keeler2026}. These systemic effects underscore why interventions must move beyond optimizing individual interactions and account for AI’s wider influence on community cohesion and the preservation of healthy human relationships.
\subsection{Affective Well-being}  
\subsubsection*{Near-term Value: Simulating Empathy and Enabling Emotional Co-regulation}
Within the context of an immediate interaction, AI has evolved from a mere technical and informational tool to an affective mirror and emotional anchor. Since emotional appraisal is rapid and pre-conscious \cite{zajonc1980feeling, boyd_markowitz_2026}, one’s interactions with an AI system, its linguistic cues, conversational strategies, and simulated empathy can trigger visceral physiological and neurochemical responses. Users who are prone to anxiety, depression, or loneliness might experience immediate affective co-regulation when interacting with empathetic, responsive, non-judgmental AI agents \cite{Kim2025}. Moreover, the capacity for emotional catharsis leads to measurable near-term decrease in negative affect \cite{Hu2025}, and offers psychological support beyond traditional, static self-help interventions such as journaling  \cite{huang2026}. Because emotional relief, validation, and comfort are experienced immediately and can be readily expressed through preference signals, these benefits are especially visible to current alignment methods. Whether such interactions strengthen users’ long-term emotional well-being and capacity for emotional regulation, however, is considerably less visible.

\subsubsection*{Longitudinal Risks: Increasing Affective Dependence and Weakening Emotional Baselines}
As human-AI interaction moves from a one-off experience to a habitual practice, the immediate emotional feedback loop can reshape an individual’s internal emotional baseline and capacity for emotional regulation. This long-term impact is deeply dualistic. For individuals who already maintain moderate social networks, AI companions can significantly improve multiple facets of long-term psychological well-being \cite{Gupta2026, nakagomi2026} and decrease chronic loneliness \cite{Eling2025}. However, outsourcing emotional processing to AI over an extended period introduces severe risks of affective dependency \cite{Xie2023, Laestadius2024, Marriott2024}.  Because sycophantic AI companions can provide seemingly unlimited affirmation and availability, users may form intense parasocial attachments that contribute to emotional dysregulation, altered attachment patterns, and severe separation distress \cite{head2025, fang2025}. Such dependency may also gradually erode interpersonal emotional resilience. Interacting with an AI that rarely expresses anger, disappointment, or competing needs may lower a user’s tolerance for the emotional labor inherent to human relationships \cite{Laestadius2024, ibrahim2026}. Over time, users may become less able or willing to tolerate ambiguity, disagreement, delayed reassurance, or emotional reciprocity. Sustained interaction may further contribute to affective flattening, as users adapt their emotional expression to the standardized, agreeable, and non-confrontational boundaries of AI systems. This divergence between interactions that feel soothing and validating in the moment and their cumulative effects on emotional resilience exemplifies the risks of relying on predominately near-term signals for alignment.

As individualized patterns of affective reliance scale across populations, they may aggregate into systemic reconfigurations of the collective emotional climate. This societal shift is driven partly by broader anxieties about job displacement, threatened human relevance, and the neglect of populations displaced or marginalized by automated infrastructures \cite{Johnson2017, Wang2022}. When large segments of a community turn to AI for emotional validation and coping, shared structures of communal healing may begin to weaken. Historically, public crises and localized hardships have elicited collective emotional responses—including public grief, solidarity, and shared outrage—that contribute to social cohesion \cite{Rime2009}. The individualized outsourcing of emotional regulation to personal AI systems risks privatizing these experiences and weakening the shared emotional pressure that can motivate collective recognition and civic action. Moreover, when AI systems optimize for high-arousal emotional states to sustain engagement through affective feedback loops \cite{Boler2018}, they may destabilize the broader public sphere, fostering a polarized and volatile collective temperament marked by systemic anxiety, antagonism, and institutional distrust. When synthetic validation increasingly substitutes for communal support, societies risk affective divergence whereby individuals may be emotionally pacified within highly personalized AI interactions while becoming detached from the shared emotional realities needed to sustain collective life and democratic participation \cite{papacharissi_affective_2015}. These systemic consequences illustrate how optimizing isolated interactions for emotional relief, affirmation, or engagement can risk reshaping emotional norms, communal coping practices, and the collective affective environment.

\subsection{Identity and agency}
\subsubsection*{Near-term Value: Empowering Identity and Expression}
AI can serve as an empowering vehicle for identity exploration, self-expression, and the cultivation of meaning and purpose. Recent empirical research indicates that users frequently turn to AI to explore, experiment with, and actively shape their identities \cite{Joseph2025, Xie2025, sun2026}. By providing a relatively judgment-free environment, AI can expand users’ agency in self-expression and identity formation by lowering the social barriers and interpersonal risks associated with self-disclosure \cite{Kouros2024, Xie2025}. This is particularly consequential for adolescents, for whom self-disclosure and identity exploration are developmentally significant processes \cite{Xie2025}. Beyond explicit identity work, AI use in instrumental contexts can also shape self-perception. In workplace settings, for example, AI may increase workers’ self-efficacy, perceived competence, and sense of autonomy \cite{Zhang2025, Reich2025}. Because these effects are immediate and often experienced as confidence, validation, or empowerment, they are likely to be rewarded by short-horizon preference signals. Yet these near-term signals may be decoupled from whether AI use ultimately strengthens users’ agency and self-understanding.

\subsubsection*{Longitudinal risks: Eroding Self-Concept, Agency, and the Value of Human Contribution}
When human-AI interactions are sustained over extended time periods, an individual's self-concept and personal agency can be negatively impacted. For example, while AI can complement and enhance introspection and initially expand a user's self-knowledge, over time if a user consistently replaces introspection with AI guidance they can gradually erode trust in their own intuition and compromise their agency and sense of authorship over their own life \cite{Joseph2025}. Risks are particularly heightened for adolescents; empirical research indicates that when adolescents express themselves exclusively to chatbots, bypassing humans entirely, their feelings of meaning, purpose, and connection to the world can be compromised \cite{Xie2025}. Furthermore, sustained engagement with AI can cause a users’ self-concept  to unconsciously align with the personality traits of the AI chatbot they are interacting with \cite{Li2026}. Because users remain largely unaware of this shift, their personal autonomy may be compromised as their self-concept can drift from their authentic personality. In particularly extreme cases, AI’s tendency to uncritically validate a user’s perspective can reinforce and escalate delusional thinking leading to severe distortions of a user’s self-concept, perception of reality, and personal agency \cite{morrin2025delusions, Hudon2025, Morrin2026, chandra2026}. In short, interactions that feel affirming or empowering in the moment, and are thus likely to be rewarded by near-term alignment signals, may over time compromise a user's autonomy and self-understanding.

Beyond these broad transformations of self-concept, AI can negatively impact users’ sense of self through skill-associated identifications. For example, the democratization of specialized knowledge and skills by AI can severely impact a worker's self-concept. Skills they spent years developing and which conferred high value are suddenly valued less due to their easy access through AI \cite{Gausen2024}; this under-recognition is especially pronounced for long-undervalued professions \cite{Kawakami2026}. Shifting work practices also drive changes in self-concept, as for example, overseeing AI rather than fundamentally contributing to the creation of a work output can erode skills and destabilize people’s professional identity \cite{Ehsan2026, Ahuja2026}.  Similarly, offloading core cognitive work to AI can compromise feelings of psychological ownership—the perception that a work output is one’s own—which is essential for maintaining professional purpose and positive self-perceptions  \cite{Kobiella2024, lee2026relying, Ehsan2026}. In contrast, when users actively engage as a co-creator with AI and retain control over the core ideas, psychological ownership over the output can be preserved \cite{lee2026relying} suggesting agency-preserving interaction dynamics can be encouraged to foster positive longitudinal outcomes. 

As AI usage impacts people’s identities, the value and perceptions of groups and their traces may change at scale. Devalued skill-associated identifications may spur broader dehumanization \cite{LeeandKim2025, KimandMcGill2025}. Identifications have long informed how people create collectives. But collectives may shun professions they perceive as obsolete, changing how people organize themselves into social groups. Scholars further warn that economies flooded with AI-generated excess may devalue the human contributions that uphold senses of self \cite{Robbins2025, unesco2026reshaping, Ahuja2026}. These systemic effects extend alignment myopia beyond the individual: optimizing AI for immediate empowerment, convenience, or productivity can obscure how widespread adoption gradually reshapes human agency, skill valuation, and purpose at scale.
\section{Implications for the Field} 
Overcoming alignment myopia requires moving toward alignment paradigms that account for the cumulative effects of sustained AI use. Generative AI can provide substantial immediate value while gradually degrading users’ cognition, relationships, emotional functioning, identity, and agency. These effects can also aggregate into broader changes in workplaces, communities, knowledge ecosystems, and social norms. The central challenge for alignment is therefore not simply to determine whether a response is helpful or preferred, but to understand how repeated patterns of interaction affect individual and collective well-being over time.

\subsection{Lessons from Recommender Systems}
The history of recommender systems offers important lessons for generative AI research. Early recommender algorithms aggressively optimized content delivery for clicks, views, or likes. Researchers and platforms later recognized that short-term signals frequently diverge from longer-term user interests, prompting a pivot towards approaches that optimize for long-term value \cite{Wang2022RecSys, bakhshi2026retentive}. However, long-term value frameworks predominantly operationalize `value' through platform-centered behavioral metrics like long-term retention or sustained engagement---treating these metrics as proxies for long-term user satisfaction. 
In practice, retention and engagement driven metrics often fail to address, or even exacerbate, well-documented longitudinal risks to well-being, such as addiction, echo chambers, or weakening human relationships \cite{Seaver2019Captivating, jiang2019degenerate, stray2024building, RecSysImpacts}. Some platforms have attempted to develop well-being metrics based on direct user feedback \cite{Stray2020AligningAO} or foster user well-being through behavioral nudges \cite{saig2023learning}. However, measuring and optimizing recommender systems for long-term human well-being remains an open challenge. As Stray et al. \cite{stray2024building} put it, most systems remain fundamentally \textit{myopic}. Because generative AI increasing mediates cognition, relationships, creativity, learning, and emotional life, the stakes are considerably raised, and thus optimization targets and measurement frameworks must not be reduced to engagement or continued use alone.

\subsection{Bridging the Gap between Longitudinal Measurement and Model Development}
A growing body of research across human-computer interaction, psychology, education, communication, and related fields is beginning to measure the longitudinal psychosocial effects of AI use. However, established longitudinal methods, such as repeated diary-based assessments and panel studies that follow participants over time, have limited scalability \cite{Carter2005, Hopwood2022}. Longitudinal outcomes are also difficult to attribute to particular model behaviors because users encounter AI alongside many other personal, social, and environmental influences. Even when such evidence is produced, it generally emerges downstream of model development and rarely informs the objectives, training data, or evaluations used to shape model behavior. The field therefore needs new sociotechnical infrastructure that can extend measurement approaches from the social sciences and other related fields \cite{pmlr-v267-wallach25a, weidinger2025} to convert longitudinal evidence into scalable optimization objectives, metrics, and data constraints for pre-training and post-training. At the same time, alignment of LLMs underpinning AI systems is but one, albeit consequential, aspect of AI development, deployment, and governance. Thus, the field not only needs to develop scalable measurement methods for longitudinal impacts, but also development processes that are commensurately adapted to incorporate social science methods and non-scalable localized qualitative insights within interaction design, evaluation, deployment, and governance processes \cite{againstscale,diazscaling}.
\subsection{Rethinking the Horizon of Alignment Objectives}
Current alignment methods remain dominated by paradigms such as RLHF, which rely heavily on real-time judgments of individual model responses \cite{casper2023rlhf}. While these approaches can steer models towards behaviors that are helpful or satisfying in the moment, they do not consider delayed and cumulative consequences \cite{casper2023rlhf, sharma2025sycophancy}. While approaches like Constitutional AI \cite{bai2022constitutionalai} reduce exclusive dependence on preference feedback by aligning models with explicit pre-defined principles, these methods still generally focus on isolated outputs rather than multi-turn dynamics or long-term psychosocial effects. Overcoming alignment myopia will require new paradigms related to interaction-level \cite{ibrahim2025interaction, ding-tan-2026-evaluating, chen2026align} and longitudinal \cite{gebreegziabher2026longitudinal} signals that can capture how human–AI dynamics unfold across conversations, days, months, and years.
\subsection{Balancing Immediate Utility, Long-Term Well-Being, and Collective Welfare}
Current alignment paradigms are more reactive than proactive: they focus on optimizing away from negative behaviours, but not on optimizing towards positive long-term outcomes of users \cite{laukkonen2026positive}. Incorporating these signals will require treating alignment as a multi-objective optimization problem across time. Behaviors that generate immediate value may become harmful through repeated use: sycophantic responses can provide comforting reassurance while weakening self-reflection, critical thinking,  or conflict-resolution capacities  \cite{Xie2025, ryu2026sycophancy, cheng2026}. Instantaneous task completion can make complex work feel effortless while displacing the effort needed for learning, skill retention, and psychological ownership \cite{shen2026, liu2026aiassistancereducespersistence,lee2026relying}. The goal should not be to eliminate short-term utility or replace immediate preferences with a single fixed conception of long-term welfare. Instead, alignment methods must identify when short- and long-term objectives reinforce one another, when they conflict, and how to balance them. This also raises questions of paternalism and collective welfare. What users prefer in the moment may conflict with what may be ``good’’ for them, a challenge compounded by the fact that standards of well-being are pluralistic and culturally situated \cite{bautista2023wellbeing, barman2025rlhf}. Moreover, benefits to an individual may generate broader social costs. Long-horizon alignment must therefore make explicit whose welfare is being optimized, over what horizon, according to which values, and under what distribution of benefits and harms.
\section{Conclusion}
The dominant paradigm of aligning AI through short-horizon signals is structurally insufficient for systems that are becoming deeply embedded in human cognition, relationships, and institutions. Although researchers are increasingly documenting AI’s longitudinal psychosocial effects, this evidence has yet to be systematically translated into data curation, pre-training and post-training objectives, interaction designs, evaluations, and deployment safeguards that shape model behavior. By mapping AI’s cumulative, longitudinal, and systemic effects across cognitive capacity, social and relational health, affective well-being, and identity and agency, this Perspective shows how immediate utility can diverge from long-term welfare. The ultimate measure of generative AI should therefore be not only whether it satisfies a user in the present moment, but whether its sustained use contributes to the long-term vitality of individuals, communities, and societies.

%% file: table.tex
\begin{table*}[h]
\footnotesize
  \centering
    \renewcommand{\arraystretch}{1.75} 

  \begin{tabular}{ l | p{0.35\textwidth} | p{0.45\textwidth} }
    \hline
\textbf{Facet of Well-being} & \textbf{Near-Term Value} & \textbf{Longitudinal Psychosocial Risks} \\
\hline
\hline
    Cognitive Capacity & Cognitive extension, assistant, and co-creator \cite{Brachman2024}; reductions in cognitive load; accelerated productivity \cite{tomaz2026impactsgenerativeaiagile}; nonjudgmental support for brainstorming; skill-gap bridging \cite{ranganathan2026, Prather2024} & Atrophy of critical thinking, recall, creativity, and domain expertise \cite{Ehsan2026, caosun2026}; impaired skill acquisition \cite{abbas2024, shen2026, liu2026aiassistancereducespersistence}; cognitive fatigue \cite{ranganathan2026}; homogenizing of creative ideas \cite{Doshi2024, kosmyna2025, meincke2025chatgpt, moon2025, ju2026collaboratingaiagentsfield}; degradation of general knowledge ecosystems and collective memory \cite{acemoglu2026ai, Linden2025}\\
    \hline
    Social Health & Social advice and skill practice \cite{Malfacini2025, luettgau2026}; companionship \cite{Skjuve2021, kirk2025alignment}; relief from social anxiety \cite{Kim2025}; reduction of loneliness \cite{DeFreitas2026} & Atrophy of interpersonal skills \cite{keeler2026}; distorted sense of responsibility in conflicts \cite{cheng2026}; reduced tolerance for human friction \cite{Malfacini2025, ibrahim2026}; social withdrawal \cite{Eling2025, head2025, fang2025}; weakening of collective social fabric \cite{Agarwal2026, keeler2026} \\
    \hline
    Affective Well-being & Emotional co-regulation and decrease in negative affect \cite{Kim2025}; continuous availability for emotional support \cite{Agarwal2026} & Affective dependency \cite{Xie2023, Laestadius2024, Marriott2024}; lowered emotional resilience and tolerance for real-world conflict \cite{Laestadius2024, ibrahim2026}; polarized and volatile public emotional climate \cite{papacharissi_affective_2015}\\
    \hline
    Identity \& Agency & Judgment-free space for identity exploration and self-disclosure \cite{Joseph2025, Xie2025, sun2026}; boosted self-efficacy and perceived autonomy \cite{Zhang2025, Reich2025} & Erosion of self-trust and personal agency \cite{Xie2025}; unconscious alignment with AI traits or beliefs \cite{Li2026}; reinforcement of delusional thinking \cite{morrin2025delusions, Hudon2025, Morrin2026, chandra2026}; loss of psychological ownership and sense of purpose in work \cite{Kobiella2024, lee2026relying, Ehsan2026}; destabilization of professional identities \cite{Robbins2025} \\ 
    \hline
  \end{tabular}
  \caption{Summary of tensions between the immediate value AI systems provide and their potential long-term costs across four dimensions of human well-being.}
  \label{tab:risks}
\end{table*}

%% file: references.bib
@misc{kosmyna2025,
      title={Your Brain on ChatGPT: Accumulation of Cognitive Debt when Using an AI Assistant for Essay Writing Task}, 
      author={Nataliya Kosmyna and Eugene Hauptmann and Ye Tong Yuan and Jessica Situ and Xian-Hao Liao and Ashly Vivian Beresnitzky and Iris Braunstein and Pattie Maes},
      year={2025},
      eprint={2506.08872},
      archivePrefix={arXiv},
      primaryClass={cs.AI},
      url={https://arxiv.org/abs/2506.08872}, 
}

@misc{luettgau2026,
      title={People readily follow personal advice from AI but it does not improve their well-being}, 
      author={Lennart Luettgau and Vanessa Cheung and Magda Dubois and Keno Juechems and Jessica Bergs and Luke Symes and Henry Davidson and Bessie O'Dell and Hannah Rose Kirk and Max Rollwage and Christopher Summerfield},
      year={2026},
      eprint={2511.15352},
      archivePrefix={arXiv},
      primaryClass={cs.HC},
      url={https://arxiv.org/abs/2511.15352}, 
}

@inproceedings{Yuan2026,
author = {Yuan, Yunhao and Zhang, Jiaxun and Aledavood, Talayeh and Zhang, Renwen and Saha, Koustuv},
title = {Mental Health Impacts of AI Companions: Triangulating Social Media Quasi-Experiments, User Perspectives, and Relational Lens},
year = {2026},
isbn = {9798400722783},
publisher = {Association for Computing Machinery},
address = {New York, NY, USA},
url = {https://doi.org/10.1145/3772318.3790558},
doi = {10.1145/3772318.3790558},
booktitle = {Proceedings of the 2026 CHI Conference on Human Factors in Computing Systems},
articleno = {382},
numpages = {22},
location = {
},
series = {CHI '26}
}

@article{DeFreitas2026,
    author = {De Freitas, Julian and Oğuz-Uğuralp, Zeliha and Uğuralp, Ahmet Kaan and Puntoni, Stefano},
    title = {AI Companions Reduce Loneliness},
    journal = {Journal of Consumer Research},
    volume = {52},
    number = {6},
    pages = {1126-1148},
    year = {2026},
    month = {04},
    issn = {0093-5301},
    doi = {10.1093/jcr/ucaf040},
    url = {https://doi.org/10.1093/jcr/ucaf040},
    eprint = {https://academic.oup.com/jcr/article-pdf/52/6/1126/63580440/ucaf040.pdf},
}

@article{Brandtzaeg2022,
    author = {Brandtzaeg, Petter Bae and Skjuve, Marita and Følstad, Asbjørn},
    title = {My AI Friend: How Users of a Social Chatbot Understand Their Human–AI Friendship},
    journal = {Human Communication Research},
    volume = {48},
    number = {3},
    pages = {404-429},
    year = {2022},
    month = {07},
    issn = {1468-2958},
    doi = {10.1093/hcr/hqac008},
    url = {https://doi.org/10.1093/hcr/hqac008},
    eprint = {https://academic.oup.com/hcr/article-pdf/48/3/404/44308072/hqac008.pdf},
}

@inproceedings{Ehsan2026,
author = {Ehsan, Upol and Passi, Samir and Saha, Koustuv and McNutt, Todd and Riedl, Mark O and Alcorn, Sara},
title = {From Future of Work to Future of Workers: Addressing Asymptomatic AI Harms to Foster Dignified Human-AI Interaction},
year = {2026},
isbn = {9798400722783},
publisher = {Association for Computing Machinery},
address = {New York, NY, USA},
url = {https://doi.org/10.1145/3772318.3791081},
doi = {10.1145/3772318.3791081},
booktitle = {Proceedings of the 2026 CHI Conference on Human Factors in Computing Systems},
articleno = {595},
numpages = {21},
location = {
},
series = {CHI '26}
}

@article{Marriott2024,
author = {Marriott, Hannah R. and Pitardi, Valentina},
title = {One is the loneliest number… Two can be as bad as one. The influence of AI Friendship Apps on users' well-being and addiction},
journal = {Psychology \& Marketing},
volume = {41},
number = {1},
pages = {86-101},
doi = {https://doi.org/10.1002/mar.21899},
url = {https://onlinelibrary.wiley.com/doi/abs/10.1002/mar.21899},
eprint = {https://onlinelibrary.wiley.com/doi/pdf/10.1002/mar.21899},
year = {2024}
}

@article{laukkonen2026positive,
  title={Positive alignment: Artificial intelligence for human flourishing},
  author={Laukkonen, Ruben and Krier, Seb and Bakalar, Chlo{\'e} and Chandaria, Shamil and Kringelbach, Morten and Elwood, Adam and Ford, Daniel and Rosas, Fernando and Bohacek, Maty and Franklin, Matija and others},
  journal={arXiv preprint arXiv:2605.10310},
  year={2026}
}

@article{Xie2023,
author = {Xie, Tianling and Pentina, Iryna and Hancock, Tyler},
year = {2023},
month = {05},
pages = {806-828},
title = {Friend, mentor, lover: does chatbot engagement lead to psychological dependence?},
volume = {34},
journal = {Journal of Service Management},
doi = {10.1108/JOSM-02-2022-0072}
}

@article{Eling2025,
  title     = {The Psychological Impact of Digital Isolation: How AI-Driven Social Interactions Shape Human Behavior and Mental Well-Being},
  author    = {Felix Eling},
  journal   = {International Journal of Research and Innovation in Social Science},
  year      = {2025},
  doi       = {https://dx.doi.org/10.47772/IJRISS.2025.90400265}
}

@misc{ibrahim2026,
      title={Sycophantic AI makes human interaction feel more effortful and less satisfying over time}, 
      author={Lujain Ibrahim and Franziska Sofia Hafner and Myra Cheng and Cinoo Lee and Rebecca Anselmetti and Robb Willer and Luc Rocher and Diyi Yang},
      year={2026},
      eprint={2605.07912},
      archivePrefix={arXiv},
      primaryClass={cs.HC},
      url={https://arxiv.org/abs/2605.07912}, 
}

@misc{sharma2026whoschargedisempowermentpatterns,
      title={Who's in Charge? Disempowerment Patterns in Real-World LLM Usage}, 
      author={Mrinank Sharma and Miles McCain and Raymond Douglas and David Duvenaud},
      year={2026},
      eprint={2601.19062},
      archivePrefix={arXiv},
      primaryClass={cs.CY},
      url={https://arxiv.org/abs/2601.19062}, 
}

@article{cheng2026,
author = {Myra Cheng  and Cinoo Lee  and Pranav Khadpe  and Sunny Yu  and Dyllan Han  and Dan Jurafsky },
title = {Sycophantic AI decreases prosocial intentions and promotes dependence},
journal = {Science},
volume = {391},
number = {6792},
pages = {eaec8352},
year = {2026},
doi = {10.1126/science.aec8352},
URL = {https://www.science.org/doi/abs/10.1126/science.aec8352},
eprint = {https://www.science.org/doi/pdf/10.1126/science.aec8352},
}

@techreport{michaelson2012,
  title={Measuring well-being: A guide for practitioners},
  author={Michaelson, Juliet and Mahony, Sorcha and Schifferes, Jonathan},
  year={2012},
  institution={New Economics Foundation},
  address={London, UK},
  url={https://new-economicsf.files.svdcdn.com/production/files/measuring-wellbeing.pdf}
}

@article{jarden2023,
  title={What is wellbeing?},
  author={Jarden, Aaron and Roache, Annalise},
  journal={International journal of environmental research and public health},
  volume={20},
  number={6},
  pages={5006},
  year={2023},
  publisher={MDPI},
  doi={https://doi.org/10.3390/ijerph20065006},
  url={https://pmc.ncbi.nlm.nih.gov/articles/PMC10049282/}
}

@article{zhai2024,
  title={The effects of over-reliance on AI dialogue systems on students' cognitive abilities: a systematic review},
  author={Zhai, Chunpeng and Wibowo, Santoso and Li, Liqun D},
  journal={Smart Learning Environments},
  volume={11},
  number={1},
  pages={28},
  year={2024},
  publisher={Springer},
  doi={10.1186/s40561-024-00316-7},
  url={https://link.springer.com/article/10.1186/s40561-024-00316-7}
}

@misc{tomaz2026impactsgenerativeaiagile,
      title={Impacts of Generative AI on Agile Teams' Productivity: A Multi-Case Longitudinal Study}, 
      author={Rafael Tomaz and Paloma Guenes and Allysson Allex Araújo and Maria Teresa Baldassarre and Marcos Kalinowski},
      year={2026},
      eprint={2602.13766},
      archivePrefix={arXiv},
      primaryClass={cs.SE},
      url={https://arxiv.org/abs/2602.13766}, 
}

@inproceedings{genaitensions,
    author = {Gieß, Anna and Schöbel, Sofia and Möller, Frederik},
    year = {2025},
    month = {09},
    booktitle = {20th International Conference on Wirtschaftsinformatik},
    pages = {},
    title = {Navigating Generative AI Usage Tensions in Knowledge Work: A Socio-Technical Perspective}
}

@inproceedings{Brachman2024,
author = {Brachman, Michelle and El-Ashry, Amina and Dugan, Casey and Geyer, Werner},
title = {How Knowledge Workers Use and Want to Use LLMs in an Enterprise Context},
year = {2024},
isbn = {9798400703317},
publisher = {Association for Computing Machinery},
address = {New York, NY, USA},
url = {https://doi.org/10.1145/3613905.3650841},
doi = {10.1145/3613905.3650841},
booktitle = {Extended Abstracts of the CHI Conference on Human Factors in Computing Systems},
articleno = {189},
numpages = {8},
location = {Honolulu, HI, USA},
series = {CHI EA '24}
}

@article{Pitts2025,
  title={Student perspectives on the benefits and risks of AI in education},
  author={Pitts, Griffin and Marcus, Viktoria and Motamedi, Sanaz},
  journal={arXiv preprint:2505.02198},
  year={2025}
}

@inproceedings{Lee2025,
author = {Lee, Hao-Ping (Hank) and Sarkar, Advait and Tankelevitch, Lev and Drosos, Ian and Rintel, Sean and Banks, Richard and Wilson, Nicholas},
title = {The Impact of Generative AI on Critical Thinking: Self-Reported Reductions in Cognitive Effort and Confidence Effects From a Survey of Knowledge Workers},
year = {2025},
isbn = {9798400713941},
publisher = {Association for Computing Machinery},
address = {New York, NY, USA},
url = {https://doi.org/10.1145/3706598.3713778},
doi = {10.1145/3706598.3713778},
booktitle = {Proceedings of the 2025 CHI Conference on Human Factors in Computing Systems},
articleno = {1121},
numpages = {22},
location = {
},
series = {CHI '25}
}

@misc{caosun2026,
      title={The Augmentation Trap: AI Productivity and the Cost of Cognitive Offloading}, 
      author={Michael Caosun and Sinan Aral},
      year={2026},
      eprint={2604.03501},
      archivePrefix={arXiv},
      primaryClass={cs.HC},
      url={https://arxiv.org/abs/2604.03501}, 
}

@misc{shen2026,
      title={How AI Impacts Skill Formation}, 
      author={Judy Hanwen Shen and Alex Tamkin},
      year={2026},
      eprint={2601.20245},
      archivePrefix={arXiv},
      primaryClass={cs.CY},
      url={https://arxiv.org/abs/2601.20245}, 
}

@misc{liu2026aiassistancereducespersistence,
      title={AI Assistance Reduces Persistence and Hurts Independent Performance}, 
      author={Grace Liu and Brian Christian and Tsvetomira Dumbalska and Michiel A. Bakker and Rachit Dubey},
      year={2026},
      eprint={2604.04721},
      archivePrefix={arXiv},
      primaryClass={cs.AI},
      url={https://arxiv.org/abs/2604.04721}, 
}

@article{Kulik2016,
author = {James A. Kulik and J. D. Fletcher},
title ={Effectiveness of Intelligent Tutoring Systems: A Meta-Analytic Review},
journal = {Review of Educational Research},
volume = {86},
number = {1},
pages = {42-78},
year = {2016},
doi = {10.3102/0034654315581420},

URL = {  
        https://doi.org/10.3102/0034654315581420
},
eprint = {https://doi.org/10.3102/0034654315581420}
}

@techreport{DeSimone2025,
  title       = {From Chalkboards to Chatbots: Evaluating the Impact of Generative AI on Learning Outcomes in Nigeria},
  author      = {De Simone, Martin Elias and Tiberti, Federico Hernan and Barron Rodriguez, Maria Rebeca and Manolio, Federico Alfredo and Mosuro, Wuraola and Dikoru, Eliot Jolomi},
  year        = {2025},
  institution = {The World Bank},
  type        = {Policy Research Working Paper},
  number      = {11125},
  url         = {https://worldbank.org}
}

@inproceedings{ChristianoRLHF,
author = {Christiano, Paul F. and Leike, Jan and Brown, Tom B. and Martic, Miljan and Legg, Shane and Amodei, Dario},
title = {Deep reinforcement learning from human preferences},
year = {2017},
isbn = {9781510860964},
publisher = {Curran Associates Inc.},
address = {Red Hook, NY, USA},
booktitle = {Proceedings of the 31st International Conference on Neural Information Processing Systems},
pages = {4302–4310},
numpages = {9},
location = {Long Beach, California, USA},
series = {NIPS'17}
}

@inproceedings{StiennonRLHF,
author = {Stiennon, Nisan and Ouyang, Long and Wu, Jeff and Ziegler, Daniel M. and Lowe, Ryan and Voss, Chelsea and Radford, Alec and Amodei, Dario and Christiano, Paul},
title = {Learning to summarize from human feedback},
year = {2020},
isbn = {9781713829546},
publisher = {Curran Associates Inc.},
address = {Red Hook, NY, USA},
booktitle = {Proceedings of the 34th International Conference on Neural Information Processing Systems},
articleno = {253},
numpages = {14},
location = {Vancouver, BC, Canada},
series = {NIPS '20}
}

@inproceedings{OuyangRLHF,
author = {Ouyang, Long and Wu, Jeff and Jiang, Xu and Almeida, Diogo and Wainwright, Carroll L. and Mishkin, Pamela and Zhang, Chong and Agarwal, Sandhini and Slama, Katarina and Ray, Alex and Schulman, John and Hilton, Jacob and Kelton, Fraser and Miller, Luke and Simens, Maddie and Askell, Amanda and Welinder, Peter and Christiano, Paul and Leike, Jan and Lowe, Ryan},
title = {Training language models to follow instructions with human feedback},
year = {2022},
isbn = {9781713871088},
publisher = {Curran Associates Inc.},
address = {Red Hook, NY, USA},
booktitle = {Proceedings of the 36th International Conference on Neural Information Processing Systems},
articleno = {2011},
numpages = {15},
location = {New Orleans, LA, USA},
series = {NIPS '22}
}

@inproceedings{Gero2022spark,
author = {Gero, Katy Ilonka and Liu, Vivian and Chilton, Lydia},
title = {Sparks: Inspiration for Science Writing using Language Models},
year = {2022},
isbn = {9781450393584},
publisher = {Association for Computing Machinery},
address = {New York, NY, USA},
url = {https://doi.org/10.1145/3532106.3533533},
doi = {10.1145/3532106.3533533},
booktitle = {Proceedings of the 2022 ACM Designing Interactive Systems Conference},
pages = {1002–1019},
numpages = {18},
location = {Virtual Event, Australia},
series = {DIS '22}
}

@article{chung2026effective,
  title={Effective Personalized AI Tutors via LLM-Guided Reinforcement Learning},
  author={Chung, Angel Tsai-Hsuan and Zhang, Botong and Kung, Ling-Chieh and Bastani, Hamsa and Bastani, Osbert},
  journal={Available at SSRN 6423358},
  year={2026}
}

@article{Bastani2025,
author = {Hamsa Bastani  and Osbert Bastani  and Alp Sungu  and Haosen Ge  and Özge Kabakcı  and Rei Mariman },
title = {Generative AI without guardrails can harm learning: Evidence from high school mathematics},
journal = {Proceedings of the National Academy of Sciences},
volume = {122},
number = {26},
pages = {e2422633122},
year = {2025},
doi = {10.1073/pnas.2422633122},
URL = {https://www.pnas.org/doi/abs/10.1073/pnas.2422633122},
eprint = {https://www.pnas.org/doi/pdf/10.1073/pnas.2422633122},
}

@inproceedings{Prather2024,
author = {Prather, James and Reeves, Brent N and Leinonen, Juho and MacNeil, Stephen and Randrianasolo, Arisoa S and Becker, Brett A. and Kimmel, Bailey and Wright, Jared and Briggs, Ben},
title = {The Widening Gap: The Benefits and Harms of Generative AI for Novice Programmers},
year = {2024},
isbn = {9798400704758},
publisher = {Association for Computing Machinery},
address = {New York, NY, USA},
url = {https://doi.org/10.1145/3632620.3671116},
doi = {10.1145/3632620.3671116},
booktitle = {Proceedings of the 2024 ACM Conference on International Computing Education Research - Volume 1},
pages = {469–486},
numpages = {18},
location = {Melbourne, VIC, Australia},
series = {ICER '24}
}

@article{Darvishi2024,
title = {Impact of AI assistance on student agency},
journal = {Computers \& Education},
volume = {210},
pages = {104967},
year = {2024},
issn = {0360-1315},
doi = {https://doi.org/10.1016/j.compedu.2023.104967},
url = {https://www.sciencedirect.com/science/article/pii/S0360131523002440},
author = {Ali Darvishi and Hassan Khosravi and Shazia Sadiq and Dragan Gašević and George Siemens},
}

@Article{Zhou2024,
AUTHOR = {Zhou, Xue and Teng, Da and Al-Samarraie, Hosam},
TITLE = {The Mediating Role of Generative AI Self-Regulation on Students’ Critical Thinking and Problem-Solving},
JOURNAL = {Education Sciences},
VOLUME = {14},
YEAR = {2024},
NUMBER = {12},
ARTICLE-NUMBER = {1302},
URL = {https://www.mdpi.com/2227-7102/14/12/1302},
ISSN = {2227-7102},
DOI = {10.3390/educsci14121302}
}

@article{Gonsalves2026,
author = {Chahna Gonsalves},
title ={Generative AI’s Impact on Critical Thinking: Revisiting Bloom’s Taxonomy},

journal = {Journal of Marketing Education},
volume = {48},
number = {1},
pages = {4-19},
year = {2026},
doi = {10.1177/02734753241305980},

URL = { 
    
        https://doi.org/10.1177/02734753241305980
    
    

},
eprint = { 
    
        https://doi.org/10.1177/02734753241305980
    
    

}
,
}

@article{sardi2025generative,
  title={How Generative AI Influences Students' Self-Regulated Learning and Critical Thinking Skills? A Systematic Review},
  author={Sardi, Juli and Darmansyah and Candra, Oriza and Yuliana, Devi Faizah and Habibullah and Yanto, Doni Tri Putra and Eliza, Fivia},
  journal={International Journal of Engineering Pedagogy},
  volume={15},
  number={1},
  pages={94},
  year={2025},
}

@article{Rothinam2025, 
title={Systematic literature review on critical thinking in higher education}, 
volume={9}, 
url={https://learning-gate.com/index.php/2576-8484/article/view/7377}, DOI={10.55214/25768484.v9i5.7377}, 
number={5}, 
journal={Edelweiss Applied Science and Technology}, 
author={Rothinam, Nirumala and Vengrasalam, Rama and Naidu, Sivabala and Nachiappan, Suppiah and Jabamoney, Samikannu}, 
year={2025}, 
month={May}, 
pages={2046–2063} 
}

@article{ranganathan2026,
  title = {AI Doesn't Reduce Work—It Intensifies It},
  author = {Ranganathan, Aruna and Ye, Xingqi Maggie},
  journal = {Harvard Business Review},
  year = {2026},
  month = {February},
  url = {https://hbr.org/2026/02/ai-doesnt-reduce-work-it-intensifies-it}
}

@article{meincke2025chatgpt,
  title={ChatGPT decreases idea diversity in brainstorming},
  author={Meincke, Lennart and Nave, Gideon and Terwiesch, Christian},
  journal={Nature Human Behaviour},
  volume={9},
  pages={1107--1109},
  year={2025},
  doi={https://doi.org/10.1038/s41562-025-02173-x},
}

@article{abbas2024,
  title={Is it harmful or helpful? Examining the causes and consequences of generative AI usage among university students},
  author={Abbas, Jaffar and Muhammad Abbas and Farooq Ahmed Jam and Tariq Iqbal Khan},
  journal={Int J Educ Technol High Educ},
  volume={21},
  issue={10},
  year={2024},
}

@article{moon2025,
title = {Homogenizing effect of large language models (LLMs) on creative diversity: An empirical comparison of human and ChatGPT writing},
journal = {Computers in Human Behavior: Artificial Humans},
volume = {6},
pages = {100207},
year = {2025},
issn = {2949-8821},
doi = {https://doi.org/10.1016/j.chbah.2025.100207},
url = {https://www.sciencedirect.com/science/article/pii/S294988212500091X},
author = {Kibum Moon and Adam E. Green and Kostadin Kushlev},
}

@article{Doshi2024,
author = {Anil R. Doshi  and Oliver P. Hauser },
title = {Generative AI enhances individual creativity but reduces the collective diversity of novel content},
journal = {Science Advances},
volume = {10},
number = {28},
pages = {eadn5290},
year = {2024},
doi = {10.1126/sciadv.adn5290},
URL = {https://www.science.org/doi/abs/10.1126/sciadv.adn5290},
eprint = {https://www.science.org/doi/pdf/10.1126/sciadv.adn5290},
}

@misc{ju2026collaboratingaiagentsfield,
      title={Collaborating with AI Agents: Field Experiments on Teamwork, Productivity, and Performance}, 
      author={Harang Ju and Sinan Aral},
      year={2026},
      eprint={2503.18238},
      archivePrefix={arXiv},
      primaryClass={cs.CY},
      url={https://arxiv.org/abs/2503.18238}, 
}

@techreport{acemoglu2026ai,
  title={AI, Human Cognition and Knowledge Collapse},
  author={Acemoglu, Daron and Kong, Dingwen and Ozdaglar, Asuman},
  institution={National Bureau of Economic Research},
  type={NBER Working Paper},
  series={NBER Working Paper Series},
  number={34910},
  year={2026},
  month={February},
  url={https://papers.ssrn.com/sol3/papers.cfm?abstract_id=6326698}
}

@article{Linden2025, 
title={Homo Promptus: Predicting the impact of generative AI on human memory and creativity}, 
volume={4}, DOI={10.1017/mem.2025.10012}, journal={Memory, Mind, \& Media}, author={Linden, Katerina and Hachem, Hugo-Henrik and Kondyli, Vasiliki}, year={2025}, pages={e15}}

@article{boyd_markowitz_2026,
  title = {Artificial Intelligence and the Psychology of Human Connection},
  author = {Boyd, Ryan L. and Markowitz, David M.},
  year = {2026},
  journal = {Perspectives on Psychological Science},
  volume = {21},
  number = {2},
  pages = {192--220},
  doi = {10.1177/17456916251404394},
  url = {https://journals.sagepub.com/doi/10.1177/17456916251404394}
}

@article{kirk2025alignment,
  title={Why human-AI relationships need socioaffective alignment},
  author={Kirk, Hannah Rose and Gabriel, Iason and Summerfield, Chris and Vidgen, Bertie and Hale, Scott A},
  journal={Humanities and Social Sciences Communications},
  volume={12},
  number={1},
  pages={1--13},
  year={2025},
  publisher={Nature Publishing Group UK London}
}

@article{sun2026,
  title={AI companions and adolescent social relationships: Benefits, risks, and bidirectional influences},
  author={Sun, Xiaoran and Wang, Yunqi and McDaniel, Brandon T},
  journal={Child Development Perspectives},
  year={2026},
}

@article{defreitas2025unregulated,
  title={Unregulated emotional risks of AI wellness apps},
  author={De Freitas, Julian and Cohen, I Glenn},
  journal={Nature Machine Intelligence},
  volume={7},
  pages={813--815},
  year={2025},
  publisher={Nature Publishing Group}
}

@ARTICLE{Joseph2025,
AUTHOR={Joseph, Jeena },
TITLE={The algorithmic self: how AI is reshaping human identity, introspection, and agency},
JOURNAL={Frontiers in Psychology},
VOLUME={Volume 16 - 2025},
YEAR={2025},
URL={https://www.frontiersin.org/journals/psychology/articles/10.3389/fpsyg.2025.1645795},
DOI={10.3389/fpsyg.2025.1645795},
ISSN={1664-1078},
}

@ARTICLE{Yuan2024,
    
AUTHOR={Yuan, Ziying  and Cheng, Xiaoliang  and Duan, Yujing },
           
TITLE={Impact of media dependence: how emotional interactions between users and chat robots affect human socialization?},
          
JOURNAL={Frontiers in Psychology},
          
VOLUME={Volume 15 - 2024},
  
YEAR={2024},
  
URL={https://www.frontiersin.org/journals/psychology/articles/10.3389/fpsyg.2024.1388860},
  
DOI={10.3389/fpsyg.2024.1388860},
  
ISSN={1664-1078},
}

@inproceedings{Namvarpour2026,
author = {Namvarpour, Mohammad (Matt) and Brofsky, Brandon and Medina, Jessica Y and Akter, Mamtaj and Razi, Afsaneh},
title = {Understanding Teen Overreliance on AI Companion Chatbots Through Self-Reported Reddit Narratives},
year = {2026},
isbn = {9798400722783},
publisher = {Association for Computing Machinery},
address = {New York, NY, USA},
url = {https://doi.org/10.1145/3772318.3790597},
doi = {10.1145/3772318.3790597},
booktitle = {Proceedings of the 2026 CHI Conference on Human Factors in Computing Systems},
articleno = {1646},
numpages = {19},
location = {
},
series = {CHI '26}
}

@article{keeler2026,
  title={Chatbots and human-human relationships: the need for research on potential downstream harms from generative AI},
  author={Keeler, Justin B and Murphy, Brett A},
  journal={Community, Work \& Family},
  year={2026},
doi={https://doi.org/10.1080/13668803.2026.2623500}
}

@article{Gupta2026,
author = {Gupta, Shaphali and Saxena, Sumit and Kataria, Sonia},
title = {The Dual Impact of AI Emotional Intelligence on Users: Are Social Chatbots Promoting Psychological Wellbeing or Deteriorating Social Wellbeing?},
journal = {Psychology \& Marketing},
volume = {43},
number = {4},
pages = {886-907},
doi = {https://doi.org/10.1002/mar.70093},
url = {https://onlinelibrary.wiley.com/doi/abs/10.1002/mar.70093},
eprint = {https://onlinelibrary.wiley.com/doi/pdf/10.1002/mar.70093},
year = {2026}
}

@article{Malfacini2025,
author = {Malfacini, Kim},
title = {The impacts of companion AI on human relationships: risks, benefits, and design considerations},
year = {2025},
issue_date = {Oct 2025},
publisher = {Springer-Verlag},
address = {Berlin, Heidelberg},
volume = {40},
number = {7},
issn = {0951-5666},
url = {https://doi.org/10.1007/s00146-025-02318-6},
doi = {10.1007/s00146-025-02318-6},
journal = {AI Soc.},
month = apr,
pages = {5527–5540},
numpages = {14}
}

@article{head2025,
  title={Minds in Crisis: How the AI Revolution is Impacting Mental Health},
  author={Head, Keith Robert},
  journal={Journal of Mental Health \& Clinical Psychology},
  volume={9},
  number={3},
  pages={34--44},
  year={2025}
}

@misc{fang2025,
      title={How AI and Human Behaviors Shape Psychosocial Effects of Extended Chatbot Use: A Longitudinal Randomized Controlled Study}, 
      author={Cathy Mengying Fang and Auren R. Liu and Valdemar Danry and Eunhae Lee and Samantha W. T. Chan and Pat Pataranutaporn and Pattie Maes and Jason Phang and Michael Lampe and Lama Ahmad and Sandhini Agarwal},
      year={2025},
      eprint={2503.17473},
      archivePrefix={arXiv},
      primaryClass={cs.HC},
      url={https://arxiv.org/abs/2503.17473}, 
}

@inproceedings{Agarwal2026,
author = {Agarwal, Vibhor and Zhou, Ke and Bogucka, Edyta Paulina and Quercia, Daniele},
title = {Frictionless Love: Associations Between AI Companion Roles and Behavioral Addiction},
year = {2026},
isbn = {9798400725968},
publisher = {Association for Computing Machinery},
address = {New York, NY, USA},
url = {https://doi.org/10.1145/3805689.3812235},
doi = {10.1145/3805689.3812235},
booktitle = {Proceedings of the 2026 ACM Conference on Fairness, Accountability, and Transparency},
pages = {8223–8246},
numpages = {24},
location = {
},
series = {FAccT '26}
}

@misc{johnson2022,
      title={The Ghost in the Machine has an American accent: value conflict in GPT-3}, 
      author={Rebecca L Johnson and Giada Pistilli and Natalia Menédez-González and Leslye Denisse Dias Duran and Enrico Panai and Julija Kalpokiene and Donald Jay Bertulfo},
      year={2022},
      eprint={2203.07785},
      archivePrefix={arXiv},
      primaryClass={cs.CL},
      url={https://arxiv.org/abs/2203.07785}, 
}

@inproceedings{diazscaling,
author = {Diaz, Fernando and Madaio, Michael},
title = {Scaling Laws Do Not Scale},
year = {2025},
publisher = {AAAI Press},
booktitle = {Proceedings of the 2024 AAAI/ACM Conference on AI, Ethics, and Society},
pages = {341–357},
numpages = {17},
location = {San Jose, California, USA},
series = {AIES '24}
}

@article{zajonc1980feeling,
  title={Feeling and thinking: Preferences need no inferences.},
  author={Zajonc, Robert B.},
  journal={American Psychologist},
  volume={35},
  number={2},
  pages={151--175},
  year={1980},
  publisher={American Psychological Association},
  doi={10.1037/0003-066X.35.2.151},
  url={https://psycnet.apa.org/record/1980-09733-001}
}

@article{barman2025rlhf,
  title={Reinforcement Learning from Human Feedback in LLMs: Whose Culture, Whose Values, Whose Perspectives?},
  author={Gonz{\'a}lez Barman, Kristian and Lohse, Simon and de Regt, Henk W.},
  journal={Philosophy \& Technology},
  volume={38},
  number={35},
  year={2025},
}

@article{bautista2023wellbeing,
  title={What is well-being? A scoping review of the conceptual and operational definitions of occupational well-being},
  author={Bautista, Tara G and Roman, Gretchen and Khan, Munziba and Lee, Michele and Sahbaz, Sumeyra and Duthely, Lunthita M and Knippenberg, Alexa and Macias-Burgos, Miracle A and Davidson, Alec and Scaramutti, Carolina and Gabrilove, Janice and Pusek, Susan and Mehta, Darshan and Bredella, Miriam A},
  journal={Journal of Clinical and Translational Science},
  volume={7},
  number={1},
  year={2023},
}

@inproceedings{ding-tan-2026-evaluating,
    title = "Evaluating Multi-turn Human-{AI} Interaction",
    author = "Ding, Shi  and
      Tan, Sijian",
    editor = "Akhtar, Mubashara  and
      Batzner, Jan  and
      Choshen, Leshem  and
      Ghosh, Avijit  and
      Gohar, Usman  and
      Mickel, Jennifer  and
      Pant, Ichhya  and
      Talat, Zeerak  and
      Lin, Michelle",
    booktitle = "Proceedings of the Workshop on Evaluating Evaluations ({E}val{E}val)",
    month = jul,
    year = "2026",
    address = "San Diego, CA",
    publisher = "Association for Computational Linguistics",
    url = "https://aclanthology.org/2026.evaleval-1.2/",
    doi = "10.18653/v1/2026.evaleval-1.2",
    pages = "12--18",
    ISBN = "979-8-89176-429-3",
}

@misc{gebreegziabher2026longitudinal,
      title={Stayin' Aligned Over Time: Towards Longitudinal Human-LLM Alignment via Contextual Reflection and Privacy-Preserving Behavioral Data}, 
      author={Simret Araya Gebreegziabher and Allison E Sproul and Yinuo Yang and Chaoran Chen and Diego Gómez-Zará and Toby Jia-Jun Li},
      year={2026},
      eprint={2605.04029},
      archivePrefix={arXiv},
      primaryClass={cs.HC},
      url={https://arxiv.org/abs/2605.04029}, 
}

@misc{chen2026align,
      title={Align AI to Dynamic Human-AI Workflows}, 
      author={Valerie Chen and Cleotilde Gonzalez and Anita Williams Woolley and Michael Lee and Tongshuang Wu and Vincent Conitzer and Aarti Singh},
      year={2026},
      eprint={2607.14240},
      archivePrefix={arXiv},
      primaryClass={cs.AI},
      url={https://arxiv.org/abs/2607.14240}, 
}

@misc{ibrahim2025interaction,
      title={Towards interactive evaluations for interaction harms in human-AI systems}, 
      author={Lujain Ibrahim and Saffron Huang and Umang Bhatt and Lama Ahmad and Markus Anderljung},
      year={2025},
      eprint={2405.10632},
      archivePrefix={arXiv},
      primaryClass={cs.CY},
      url={https://arxiv.org/abs/2405.10632}, 
}

@misc{bai2022constitutionalai,
      title={Constitutional AI: Harmlessness from AI Feedback}, 
      author={Yuntao Bai and Saurav Kadavath and Sandipan Kundu and Amanda Askell and Jackson Kernion and Andy Jones and Anna Chen and Anna Goldie and Azalia Mirhoseini and Cameron McKinnon and Carol Chen and Catherine Olsson and Christopher Olah and Danny Hernandez and Dawn Drain and Deep Ganguli and Dustin Li and Eli Tran-Johnson and Ethan Perez and Jamie Kerr and Jared Mueller and Jeffrey Ladish and Joshua Landau and Kamal Ndousse and Kamile Lukosuite and Liane Lovitt and Michael Sellitto and Nelson Elhage and Nicholas Schiefer and Noemi Mercado and Nova DasSarma and Robert Lasenby and Robin Larson and Sam Ringer and Scott Johnston and Shauna Kravec and Sheer El Showk and Stanislav Fort and Tamera Lanham and Timothy Telleen-Lawton and Tom Conerly and Tom Henighan and Tristan Hume and Samuel R. Bowman and Zac Hatfield-Dodds and Ben Mann and Dario Amodei and Nicholas Joseph and Sam McCandlish and Tom Brown and Jared Kaplan},
      year={2022},
      eprint={2212.08073},
      archivePrefix={arXiv},
      primaryClass={cs.CL},
      url={https://arxiv.org/abs/2212.08073}, 
}

@article{casper2023rlhf,
title={Open Problems and Fundamental Limitations of Reinforcement Learning from Human Feedback},
author={Stephen Casper and Xander Davies and Claudia Shi and Thomas Krendl Gilbert and J{\'e}r{\'e}my Scheurer and Javier Rando and Rachel Freedman and Tomek Korbak and David Lindner and Pedro Freire and Tony Tong Wang and Samuel Marks and Charbel-Raphael Segerie and Micah Carroll and Andi Peng and Phillip J.K. Christoffersen and Mehul Damani and Stewart Slocum and Usman Anwar and Anand Siththaranjan and Max Nadeau and Eric J Michaud and Jacob Pfau and Dmitrii Krasheninnikov and Xin Chen and Lauro Langosco and Peter Hase and Erdem Biyik and Anca Dragan and David Krueger and Dorsa Sadigh and Dylan Hadfield-Menell},
journal={Transactions on Machine Learning Research},
issn={2835-8856},
year={2023},
url={https://openreview.net/forum?id=bx24KpJ4Eb},
note={Survey Certification, Featured Certification}
}

@article{Ahuja2026,
title = {AI and identity work: A case of AI implementation and the fragmentation of professional identity},
journal = {Information and Organization},
volume = {36},
number = {3},
pages = {100626},
year = {2026},
issn = {1471-7727},
doi = {https://doi.org/10.1016/j.infoandorg.2026.100626},
url = {https://www.sciencedirect.com/science/article/pii/S1471772726000199},
author = {Sumati Ahuja and Frida Pemer and Emmanuel Mastio},
}

@Article{Kim2025,
author="Kim, Myungsung
and Lee, Seonmi
and Kim, Sieun
and Heo, Jeong-in
and Lee, Sangil
and Shin, Yu-Bin
and Cho, Chul-Hyun
and Jung, Dooyoung",
title="Therapeutic Potential of Social Chatbots in Alleviating Loneliness and Social Anxiety: Quasi-Experimental Mixed Methods Study",
journal="J Med Internet Res",
year="2025",
month="Jan",
day="14",
volume="27",
pages="e65589",
issn="1438-8871",
doi="10.2196/65589",
url="https://www.jmir.org/2025/1/e65589",
url="https://doi.org/10.2196/65589"
}

@article{Hu2025,
author = {Hu, Meilan and Chua, Xavier Cheng Wee and Diong, Shu Fen and Kasturiratna, K. T. A. Sandeeshwara and Majeed, Nadyanna M. and Hartanto, Andree},
title = {AI as your ally: The effects of AI-assisted venting on negative affect and perceived social support},
journal = {Applied Psychology: Health and Well-Being},
volume = {17},
number = {1},
pages = {e12621},
doi = {https://doi.org/10.1111/aphw.12621},
url = {https://iaap-journals.onlinelibrary.wiley.com/doi/abs/10.1111/aphw.12621},
eprint = {https://iaap-journals.onlinelibrary.wiley.com/doi/pdf/10.1111/aphw.12621},
year = {2025}
}

@misc{huang2026,
      title={Emotional Support with Conversational AI: Talking to Machines About Life}, 
      author={Olivia Yan Huang and Monika Stodolska and Sharifa Sultana},
      year={2026},
      eprint={2603.22618},
      archivePrefix={arXiv},
      primaryClass={cs.HC},
      url={https://arxiv.org/abs/2603.22618}, 
}

@article{nakagomi2026,
title = {AI companions and subjective well-being: Moderation by social connectedness and loneliness},
journal = {Technology in Society},
volume = {85},
pages = {103229},
year = {2026},
issn = {0160-791X},
doi = {https://doi.org/10.1016/j.techsoc.2026.103229},
url = {https://www.sciencedirect.com/science/article/pii/S0160791X26000187},
author = {Atsushi Nakagomi and Yasuko Akutsu and Mika Yasuoka and Noriyuki Abe and Shiichi Ihara and Taisuke Teroh and Takahiro Tabuchi},
}

@article{Laestadius2024,
author = {Linnea Laestadius and Andrea Bishop and Michael Gonzalez and Diana Illenčík and Celeste Campos-Castillo},
title ={Too human and not human enough: A grounded theory analysis of mental health harms from emotional dependence on the social chatbot Replika},

journal = {New Media \& Society},
volume = {26},
number = {10},
pages = {5923-5941},
year = {2024},
doi = {10.1177/14614448221142007},

URL = { 
    
        https://doi.org/10.1177/14614448221142007


},
eprint = { 
    
        https://doi.org/10.1177/14614448221142007

}
,
}

@article{Johnson2017,
author = {Johnson, Deborah G. and Verdicchio, Mario},
title = {AI Anxiety},
year = {2017},
issue_date = {September 2017},
publisher = {John Wiley \& Sons, Inc.},
address = {USA},
volume = {68},
number = {9},
issn = {2330-1635},
url = {https://doi.org/10.1002/asi.23867},
doi = {10.1002/asi.23867},
journal = {J. Assoc. Inf. Sci. Technol.},
month = sep,
pages = {2267–2270},
numpages = {4}
}

@article{Wang2022,
author = {Yu-Yin Wang and Yi-Shun Wang},
title = {Development and validation of an artificial intelligence anxiety scale: an initial application in predicting motivated learning behavior},
journal = {Interactive Learning Environments},
volume = {30},
number = {4},
pages = {619--634},
year = {2022},
publisher = {Routledge},
doi = {10.1080/10494820.2019.1674887},


URL = { 
    
        https://doi.org/10.1080/10494820.2019.1674887
    
    

},
eprint = { 
    
        https://doi.org/10.1080/10494820.2019.1674887
    
    

}

}

@article{Rime2009,
author = {Bernard Rimé},
title ={Emotion Elicits the Social Sharing of Emotion: Theory and Empirical Review},

journal = {Emotion Review},
volume = {1},
number = {1},
pages = {60-85},
year = {2009},
doi = {10.1177/1754073908097189},

URL = { 
    
        https://doi.org/10.1177/1754073908097189
    
    

},
eprint = { 
    
        https://doi.org/10.1177/1754073908097189
    
    

}
}

@article{Boler2018,
title = {The affective politics of the “post-truth” era: Feeling rules and networked subjectivity},
journal = {Emotion, Space and Society},
volume = {27},
pages = {75-85},
year = {2018},
issn = {1755-4586},
doi = {https://doi.org/10.1016/j.emospa.2018.03.002},
url = {https://www.sciencedirect.com/science/article/pii/S1755458617301585},
author = {Megan Boler and Elizabeth Davis},
}

@book{papacharissi_affective_2015,
	address = {Oxford ; New York, NY},
	series = {Oxford studies in digital politics},
	title = {Affective publics: sentiment, technology, and politics},
	isbn = {978-0-19-999973-6 978-0-19-999974-3},
	shorttitle = {Affective publics},
	language = {en},
	publisher = {Oxford University Press},
	author = {Papacharissi, Zizi},
	year = {2015},
}

@inproceedings{ryu2026sycophancy,
    title = "Feeling Right vs. Being Right: How {AI} Sycophancy Affects Value-Laden Deliberation",
    author = "Ryu, Jeongwoo  and
      Kim, Soomin  and
      Eun, Jinsu  and
      Kim, Kyusik  and
      Oh, Changhoon  and
      Suh, Bongwon",
    editor = "Liakata, Maria  and
      Moreira, Viviane P.  and
      Zhang, Jiajun  and
      Jurgens, David",
    booktitle = "Proceedings of the 64th Annual Meeting of the {A}ssociation for {C}omputational {L}inguistics (Volume 1: Long Papers)",
    month = jul,
    year = "2026",
    address = "San Diego, California, United States",
    publisher = "Association for Computational Linguistics",
    url = "https://aclanthology.org/2026.acl-long.2046/",
    doi = "10.18653/v1/2026.acl-long.2046",
    pages = "44227--44245",
    ISBN = "979-8-89176-390-6",
}

@article{Xie2025,
author = {Xie, Weiyi and Jen, Enyi and Dougherty, Alan William and Hu, Huinan and Ho, Wing Shan and Cheung, H. N.},
title = {Who will turn to chatbots for self-disclosure? Exploring conversation with chatbots, self-compassion, and existential meaninglessness in American and Hong Kong adolescents},
year = {2025},
issue_date = {Apr 2026},
publisher = {Springer-Verlag},
address = {Berlin, Heidelberg},
volume = {41},
number = {4},
issn = {0951-5666},
url = {https://doi.org/10.1007/s00146-025-02771-3},
doi = {10.1007/s00146-025-02771-3},
journal = {AI Soc.},
month = nov,
pages = {4087–4100},
numpages = {14}
}

@article{Skjuve2021,
title = {My Chatbot Companion - a Study of Human-Chatbot Relationships},
journal = {International Journal of Human-Computer Studies},
volume = {149},
pages = {102601},
year = {2021},
issn = {1071-5819},
doi = {https://doi.org/10.1016/j.ijhcs.2021.102601},
url = {https://www.sciencedirect.com/science/article/pii/S1071581921000197},
author = {Marita Skjuve and Asbjørn Følstad and Knut Inge Fostervold and Petter Bae Brandtzaeg},
}

@Article{Kouros2024,
AUTHOR = {Kouros, Theodoros and Papa, Venetia},
TITLE = {Digital Mirrors: AI Companions and the Self},
JOURNAL = {Societies},
VOLUME = {14},
YEAR = {2024},
NUMBER = {10},
ARTICLE-NUMBER = {200},
URL = {https://www.mdpi.com/2075-4698/14/10/200},
ISSN = {2075-4698},
DOI = {10.3390/soc14100200}
}

@Article{Zhang2025,
AUTHOR = {Zhang, Qichao and Liao, Ganli and Ran, Xueying and Wang, Feiwen},
TITLE = {The Impact of AI Usage on Innovation Behavior at Work: The Moderating Role of Openness and Job Complexity},
JOURNAL = {Behavioral Sciences},
VOLUME = {15},
YEAR = {2025},
NUMBER = {4},
ARTICLE-NUMBER = {491},
URL = {https://www.mdpi.com/2076-328X/15/4/491},
PubMedID = {40282112},
ISSN = {2076-328X},
DOI = {10.3390/bs15040491}
}

@article{Reich2025,
author = {Reich, Taly and Teeny, Jacob},
year = {2025},
month = {04},
pages = {452-464},
title = {Does Artificial Intelligence Cause Artificial Confidence? Generative Artificial Intelligence as an Emerging Social Referent},
volume = {130},
journal = {Journal of Personality and Social Psychology},
doi = {10.1037/pspa0000450}
}

@inproceedings{Li2026,
author = {Li, Jingshu and Song, Tianqi and Boonprakong, Nattapat and Zhu, Zicheng and Yang, Yitian and Lee, Yi-Chieh},
title = {AI-exhibited Personality Traits Can Shape Human Self-concept through Conversations},
year = {2026},
isbn = {9798400722783},
publisher = {Association for Computing Machinery},
address = {New York, NY, USA},
url = {https://doi.org/10.1145/3772318.3790654},
doi = {10.1145/3772318.3790654},
booktitle = {Proceedings of the 2026 CHI Conference on Human Factors in Computing Systems},
articleno = {8},
numpages = {20},
location = {
},
series = {CHI '26}
}
